\documentclass[letterpaper, 10 pt, conference]{ieeeconf}  

\IEEEoverridecommandlockouts                              

\usepackage{amsmath} 
\usepackage{amssymb}  
\usepackage{algorithm,algpseudocode}
\usepackage{hyperref}
\usepackage[capitalize]{cleveref}
\usepackage{xspace}
\usepackage{graphicx}
\usepackage{grfext}
\usepackage{bbding}

\PrependGraphicsExtensions*{.pdf,.png,.jpg}

\makeatletter
\DeclareRobustCommand\onedot{\futurelet\@let@token\@onedot}
\def\@onedot{\ifx\@let@token.\else.\null\fi\xspace}

\def\etal{\emph{et al}\onedot}
\makeatother
\usepackage{adjustbox}
\usepackage{multirow}
\usepackage{booktabs}
\usepackage{authblk}
\usepackage{cite}
\title{\LARGE \bf
RGA-Net: A Vision Enhancement Framework for Robotic Surgical Systems Using Reciprocal Attention Mechanisms
\thanks{* Corresponding author (email: xuhangc@hzu.edu.cn).}\thanks{This work was supported in part by the Science and Technology Development Fund, Macau SAR, under Grant 0193/2023/RIA3 and 0079/2025/AFJ, and the University of Macau under Grant MYRG-GRG2024-00065-FST-UMDF, in part by Guangdong Basic and Applied Basic Research Foundation (Grant No. 2024A1515140010).} }

\author[1,2]{Quanjun Li}
\author[1,2]{Weixuan Li}
\author[1]{Han Xia}
\author[1]{Junhua Zhou}
\author[2]{Chi-Man Pun}
\author[2,3*]{Xuhang Chen}
\affil[1]{School of Advanced Manufacturing, Guangdong University of Technology}
\affil[2]{Faculty of Science and Technology, University of Macau}
\affil[3]{School of Computer Science and Engineering, Huizhou University\authorcr \href{https://github.com/QuincyQAQ/RGA-Net}{https://github.com/QuincyQAQ/RGA-Net}}

\begin{document}

\maketitle
\thispagestyle{empty}
\pagestyle{empty}

\begin{abstract}
Robotic surgical systems rely heavily on high-quality visual feedback for precise teleoperation; yet, surgical smoke from energy-based devices significantly degrades endoscopic video feeds, compromising the human-robot interface and surgical outcomes. This paper presents RGA-Net (Reciprocal Gating and Attention-fusion Network), a novel deep learning framework specifically designed for smoke removal in robotic surgery workflows. Our approach addresses the unique challenges of surgical smoke—including dense, non-homogeneous distribution and complex light scattering—through a hierarchical encoder-decoder architecture featuring two key innovations: (1) a Dual-Stream Hybrid Attention (DHA) module that combines shifted window attention with frequency-domain processing to capture both local surgical details and global illumination changes, and (2) an Axis-Decomposed Attention (ADA) module that efficiently processes multi-scale features through factorized attention mechanisms. These components are connected via reciprocal cross-gating blocks that enable bidirectional feature modulation between encoder and decoder pathways. Extensive experiments on the DesmokeData and LSD3K surgical datasets demonstrate that RGA-Net achieves superior performance in restoring visual clarity suitable for robotic surgery integration. Our method enhances the surgeon-robot interface by providing consistently clear visualization, laying a technical foundation for alleviating surgeons' cognitive burden, optimizing operation workflows, and reducing iatrogenic injury risks in minimally invasive procedures. These practical benefits could be further validated through future clinical trials involving surgeon usability assessments. The proposed framework represents a significant step toward more reliable and safer robotic surgical systems through computational vision enhancement.

\end{abstract}

\bstctlcite{IEEEexample:BSTcontrol}
\section{INTRODUCTION}
Robotic surgery, utilizing teleoperated systems like the da Vinci Surgical System, has marked a significant advancement in minimally invasive procedures by offering surgeons enhanced precision and dexterity.  However, the effectiveness of these sophisticated robotic platforms is fundamentally dependent on the quality of the visual feedback provided to the surgeon. A persistent challenge in these procedures is the generation of surgical smoke from energy-based devices, which degrades the endoscopic video feed---often the sole sensory input for the operator. This obscured surgical field can increase the surgeon's cognitive load, prolong operation times, and elevate the risk of iatrogenic injury, thereby limiting the full potential of the robotic system~\cite{pan2022desmoke}. 

Recent advances in medical image analysis have demonstrated that deep learning methods can effectively extract clinically meaningful information from complex surgical images, particularly in critical tasks such as anatomical structure delineation and precise organ segmentation. These successes highlight the potential of computational vision techniques to enhance surgical perception. However, the practical utility of these downstream analytical tasks, as well as the effectiveness of secure data transmission frameworks in safeguarding high-value clinical information, is strictly predicated on the availability of high-quality, artifact-free visual data. While hardware-based evacuation systems exist, they can be inefficient and fail to restore digital clarity in real-time. Consequently, computational image enhancement~\cite{cxh1,cxh3,cxh5} to digitally remove smoke presents a highly promising alternative for providing a consistently clear view, ensuring both human-in-the-loop safety and the robustness of automated clinical decision-making.

The task of removing smoke or haze from an image has been a long-standing problem in computer vision. Early methods relied on handcrafted statistical priors, most notably the Dark Channel Prior (DCP)~\cite{he2010single}, which assumes that most local patches in haze-free images contain low-intensity pixels in at least one color channel. While effective in outdoor scenes, such priors often fail in the unique environment of surgical scenes. With the rise of deep learning, data-driven methods have become the dominant paradigm. Architectures have evolved from early Convolutional Neural Networks (CNNs) like DehazeNet~\cite{cai2016dehazenet} and the reformulated AOD-Net~\cite{li2017aod} to more complex designs incorporating multi-scale feature fusion and attention, such as FFA-Net~\cite{qin2020ffa}. More recently, Vision Transformers (ViTs)~\cite{dosovitskiy2020vit} and their variants, like DehazeFormer~\cite{song2023vision}, have demonstrated exceptional performance.

Despite these advancements, surgical smoke removal presents distinct challenges. Unlike atmospheric haze, surgical smoke is often dense, non-homogeneously distributed, and composed of particles that cause complex light scattering, fundamentally altering scene illumination~\cite{xia2024new}. Furthermore, the development of robust, supervised learning models has been severely hampered by a critical lack of paired \textit{in vivo} datasets containing synchronous smoky and smoke-free views of the same surgical scene~\cite{pan2022desmoke,xia2024new}. This scarcity has forced reliance on synthetic data, which may not accurately model real surgical conditions, or unpaired training schemes that can lead to artifacts.

To overcome these limitations, we propose the Reciprocal Gating and Attention-fusion Network (RGA-Net), a novel deep learning architecture specifically designed to be integrated into robotic surgery workflows to address the visibility challenge. Our model adopts a U-Net structure but integrates several innovative components to robustly handle the complex characteristics of surgical smoke. The core of our network lies in two new hybrid attention modules that work in concert within an encoder-decoder framework, connected by reciprocal gating blocks that refine and fuse features across scales.

The main contributions of this work are summarized as follows:
\begin{itemize}
    \item We propose RGA-Net, a comprehensive encoder-decoder network that leverages reciprocal feature gating and multi-level attention mechanisms for effective and robust surgical smoke removal.
    \item We introduce the ADA module, a novel block that efficiently processes features by factorizing attention along block and grid axes to capture both local details and global context.
    \item We design the DHA module, which integrates spatial attention with a frequency-domain branch to learn rich feature representations, enhancing texture and structure restoration.
\end{itemize}

\section{RELATED WORK}
\subsection{Image Dehazing}
The removal of haze from images has been extensively studied. Early approaches relied on handcrafted priors. Among the most influential is the Dark Channel Prior (DCP)~\cite{he2010single}, which leverages the statistical observation that haze-free outdoor images tend to have very low intensity in at least one color channel. While effective, such prior-based methods can falter when their underlying assumptions are not met.

With the advent of deep learning, data-driven methods have become dominant. Early convolutional neural network (CNN) models like DehazeNet~\cite{cai2016dehazenet} learned to map hazy images to their clear counterparts directly. AOD-Net~\cite{li2017aod} proposed a lightweight network by reformulating the atmospheric scattering model. Subsequent works introduced more sophisticated architectures to improve performance, such as the multi-scale GridDehazeNet~\cite{liu2019griddehazenet} and the feature fusion attention mechanism in FFA-Net~\cite{qin2020ffa}. To create more robust models, AECR-Net~\cite{wu2021contrastive} introduced contrastive learning to regularize the feature space.

More recently, Vision Transformers (ViT)~\cite{dosovitskiy2020vit} have demonstrated remarkable success. Dehamer~\cite{guo2022image} and DehazeFormer~\cite{song2023vision} were among the first to adapt the transformer architecture for image dehazing, showcasing its power in modeling long-range dependencies. Concurrently, other novel approaches have emerged. MITNet~\cite{shen2023mutual} utilizes a mutual information-driven triple interaction network, DIACMPN~\cite{zhang2024depth} incorporates depth information for collaborative promotion, and SFSNiD~\cite{cong2024semi} explores semi-supervised learning for the challenging nighttime dehazing scenario. Our model builds upon the success of these transformer-based methods but introduces a novel attention and feature fusion mechanism.

\subsection{Surgical Smoke Removal}
While general-purpose dehazing algorithms can be applied to surgical scenes, the unique properties of surgical smoke---which is often dense, non-homogeneously distributed, and mixed with other visual artifacts---present distinct challenges that necessitate specialized methods.

Early deep learning-based approaches began adapting network architectures for this specific task. For instance, DesmokeNet~\cite{bolkar2018deep} introduced a deep learning framework for removing smoke from minimally invasive surgery videos, one of the initial efforts in this domain. Building on this, Wang~\etal~\cite{wang2019multiscale} proposed a multiscale architecture to better handle the varying density and distribution of smoke particles.

A significant challenge in this field is the scarcity of paired smoky and smoke-free training data. To address this, many methods have explored generative and unpaired learning techniques. Desmoke-CycleGAN~\cite{hu2021cycle} utilized cycle-consistent adversarial networks to perform image-to-image translation without requiring paired images. Similarly, Salazar~\etal~\cite{salazar2020desmoking} proposed a translation framework guided by an embedded dark channel, Sidorov~\etal~\cite{sidorov2020generative} explored other generative models, and the more recent Desmoke-LAP~\cite{pan2022desmoke} introduced an improved unpaired translation method specifically tailored for surgical desmoking.

More advanced architectures have continued to evolve. For example, MSBDN~\cite{dong2020multi}, a multi-scale boosted dehazing network, was successfully adapted to the surgical context. More recently, the Progressive Frequency-aware Network (PFAN)~\cite{zhang2023progressive} was developed to progressively remove smoke by leveraging information in the frequency domain, which helps in restoring fine details.

These works highlight a clear trend toward creating domain-specific solutions with increasingly sophisticated architectures. Our proposed RGA-Net contributes to this specialized field by designing a novel network architecture with reciprocal gating and hybrid attention mechanisms, specifically engineered for robust and efficient smoke removal in challenging surgical environments.

\section{METHODOLOGY}

Our proposed Reciprocal Gating and Attention-fusion Network (RGA-Net) employs a hierarchical encoder-decoder architecture with symmetric multi-scale processing. The network progressively encodes input features through downsampling stages, processes them in a latent space, and then reconstructs the output through upsampling stages with cross-scale feature fusion. The architecture leverages two core attention mechanisms: a dual-stream hybrid attention mechanism in the encoder and an axis-decomposed attention mechanism in the decoder and latent space.

\begin{figure*}[ht]
  \centering
  \includegraphics[width=\linewidth]{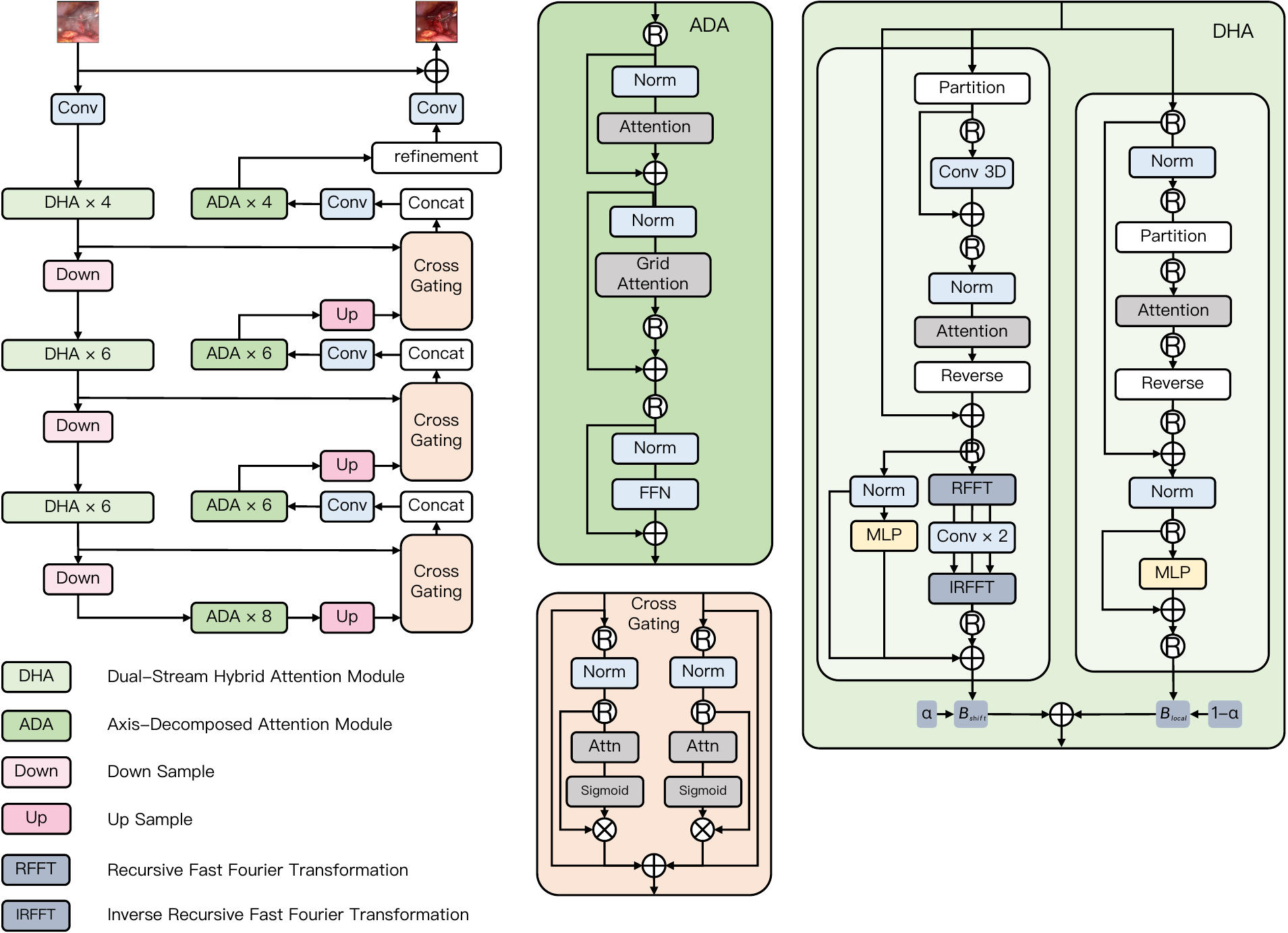}
  \caption{Overview of RGA-Net.}
  \label{fig:uie}
\end{figure*}

\subsection{Overall Architecture}

Given an input image $I \in \mathbb{R}^{H \times W \times 3}$, the network first embeds it into a feature space through a convolutional layer:
\begin{equation}
    F_0 = \text{Conv}(I) \in \mathbb{R}^{H \times W \times C},
\end{equation}
where $C$ is the base feature dimension. The encoder then processes features through $L$ levels, with each level $l$ containing multiple transformer blocks followed by downsampling:
\begin{equation}
    F_l = \text{Down}_{l-1 \rightarrow l}(\mathcal{T}_l^{\text{enc}}(F_{l-1})), \quad l = 1, 2, ..., L,
\end{equation}
where $\mathcal{T}_l^{\text{enc}}$ represents a sequence of transformer blocks and $\text{Down}$ performs spatial downsampling while increasing channel dimensions.

\subsection{Dual-Stream Hybrid Attention (DHA) Module}

The encoder employs a dual-stream hybrid attention module that processes features through two complementary pathways. For an input feature $X \in \mathbb{R}^{B \times C \times H \times W}$, the module computes:
\begin{equation}
    \text{DHA}(X) = \alpha \cdot \mathcal{B}_{\text{shift}}(X) + (1 - \alpha) \cdot \mathcal{B}_{\text{local}}(X),
\end{equation}
where $\alpha \in [0, 1]$ is a learnable parameter that balances the contributions of two attention mechanisms.

\subsubsection{Shifted Window Attention with Spectral Enhancement}

The first branch $\mathcal{B}_{\text{shift}}$ employs a shifted window attention mechanism augmented with spectral processing. The input is first partitioned into non-overlapping windows of size $M \times N$, following the Swin Transformer paradigm~\cite{liu2021swin}:
\begin{equation}
    X_{\text{windows}} = \text{Partition}(X, M, N) \in \mathbb{R}^{B \cdot \frac{HW}{MN} \times MN \times C}.
\end{equation}

For shifted attention, windows are cyclically shifted by $(\lfloor M/2 \rfloor, \lfloor N/2 \rfloor)$ in alternating blocks. Within each window, multi-head self-attention is computed with relative position bias~\cite{liu2021swin}:
\begin{equation}
    \text{Attention}(Q, K, V) = \text{Softmax}\left(\frac{QK^T}{\sqrt{d}} + B_{\text{rel}}\right)V,
\end{equation}
where $B_{\text{rel}}$ encodes learnable relative position biases between tokens within a window.

Uniquely, this branch incorporates a parallel spectral processing path inspired by frequency domain methods~\cite{ji2023single}. The feature map undergoes Fourier transformation and convolutional filtering:
\begin{equation}
    Y_{\text{spectral}} = \mathcal{F}^{-1}\left(\text{Conv}_{\mathbb{C}}(\mathcal{F}(X))\right),
\end{equation}
where $\mathcal{F}$ denotes the 2D FFT, $\mathcal{F}^{-1}$ is its inverse, and $\text{Conv}_{\mathbb{C}}$ operates on complex-valued frequency representations. The spectral branch output is added to the spatial attention output, enabling global frequency-domain processing alongside local spatial attention.

\subsubsection{Local Window Attention}
The second branch $\mathcal{B}_{\text{local}}$ implements standard window-based attention without shifting, providing stable local feature extraction~\cite{wang2022uformer}. Windows are partitioned similarly but without cyclic shifts, and attention is computed within each window using the same formulation but with different position embeddings.

\subsection{Axis-Decomposed Attention (ADA) Module}
The decoder and latent space employ axis-decomposed attention that factorizes the attention computation along different spatial axes, inspired by multi-axis vision transformers~\cite{tu2022maxim,tu2022maxvit}. For an input $X$, the module sequentially applies two types of attention:

\subsubsection{Block Attention}
Features are first partitioned into non-overlapping blocks of size $P \times Q$:
\begin{equation}
    X_{\text{blocks}} = \text{BlockPartition}(X, P, Q).
\end{equation}

Self-attention is computed within each block independently, capturing fine-grained local interactions following the blocked attention pattern in~\cite{tu2022maxvit}.

\subsubsection{Grid Attention}
Subsequently, features are repartitioned into a coarser grid structure. Instead of dividing into many small blocks, the feature map is divided into $G_h \times G_w$ grids:
\begin{equation}
    X_{\text{grids}} = \text{GridPartition}(X, G_h, G_w),
\end{equation}
where each grid cell contains $(H/G_h) \times (W/G_w)$ spatial positions. Attention is computed across positions within each grid, capturing longer-range dependencies~\cite{tu2022maxim}.

The complete ADA operation is:
\begin{align}
    X_1 &= X + \text{BlockAttn}(\text{LN}(X)), \\
    X_2 &= X_1 + \text{GridAttn}(\text{LN}(X_1)), \\
    X_{\text{out}} &= X_2 + \text{FFN}(\text{LN}(X_2)),
\end{align}
where the feed-forward network (FFN) includes depth-wise convolutions for spatial mixing, following the design in Restormer~\cite{zamir2022restormer}.

\subsection{Cross-Gating for Multi-Scale Feature Fusion}

Skip connections between encoder and decoder employ a cross-gating mechanism that enables bidirectional feature modulation, adapted from MAXIM~\cite{tu2022maxim}. Given encoder features $X_{\text{enc}}$ and decoder features $Y_{\text{dec}}$ at corresponding scales, the mechanism first aligns their dimensions:
\begin{equation}
    \hat{Y}_{\text{dec}} = \begin{cases}
        \text{ConvTranspose}(Y_{\text{dec}}) & \text{if upsampling required} \\
        \text{Conv}(Y_{\text{dec}}) & \text{otherwise}
    \end{cases}.
\end{equation}

Gating signals are generated through the ADA module:
\begin{align}
    G_x &= \sigma(\text{ADA}(\text{LN}(X_{\text{enc}}))), \\
    G_y &= \sigma(\text{ADA}(\text{LN}(\hat{Y}_{\text{dec}}))),
\end{align}
where $\sigma$ denotes the sigmoid activation.

The cross-gating operation performs reciprocal modulation:
\begin{align}
    X'_{\text{enc}} &= X_{\text{enc}} \odot G_y, \\
    Y'_{\text{dec}} &= \hat{Y}_{\text{dec}} \odot G_x,
\end{align}
where $\odot$ represents element-wise multiplication. The final fused output combines all pathways:
\begin{equation}
    X_{\text{fused}} = X'_{\text{enc}} + Y'_{\text{dec}} + X_{\text{enc}} + \hat{Y}_{\text{dec}}.
\end{equation}

This bidirectional gating allows encoder and decoder features to selectively enhance or suppress information in each other, leading to more effective multi-scale fusion than simple concatenation or addition.

\subsection{Objective Function}
To train the RGA-Net, we employ a composite objective function that combines two distinct loss terms to ensure a comprehensive restoration of the image. The network is optimized in an end-to-end manner by minimizing a weighted sum of a pixel-wise loss and a perceptual loss. Let $\hat{I}$ be the desmoked image produced by the network and $I$ be the corresponding ground-truth clear image. Our total objective function, $\mathcal{L}_{total}$, is defined as:
\begin{equation}
\mathcal{L}_{total} = \mathcal{L}_{2}(\hat{I}, I) + \lambda \cdot \mathcal{L}_{SSIM},
\end{equation}
where $\mathcal{L}_{2}$ denotes the L2 loss (Mean Squared Error), which penalizes the pixel-level differences between the predicted and ground-truth images. $\mathcal{L}_{SSIM}$ is based on the Structural Similarity Index Measure (SSIM)~\cite{wang2004image}, which evaluates perceptual similarity by considering luminance, contrast, and structure. By minimizing $\mathcal{L}_{SSIM}$, we encourage the network to preserve high-frequency structural details that are critical for visual quality. Based on empirical validation, the weighting parameter $\lambda$ is set to 0.2 to balance the contributions of the two loss components.

\section{EXPERIMENTS}

\begin{figure*}[htbp]
    \centering
    \setlength{\tabcolsep}{1pt} 
    \renewcommand{\arraystretch}{0.9} 
    \begin{tabular}{ccccccccc}


        \begin{minipage}[b]{0.11\linewidth}\centering
            \includegraphics[width=\linewidth,height=\linewidth]{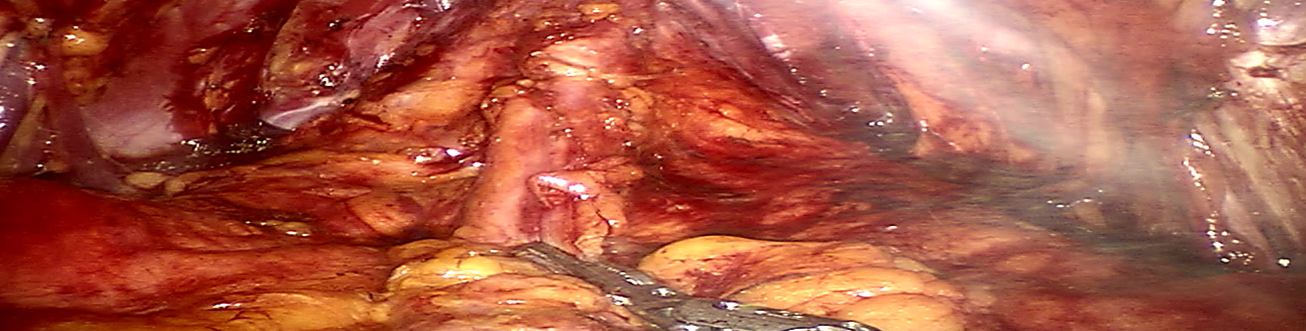}
        \end{minipage} &
        \begin{minipage}[b]{0.11\linewidth}\centering
            \includegraphics[width=\linewidth,height=\linewidth]{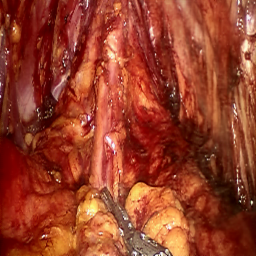} 
        \end{minipage} &
        \begin{minipage}[b]{0.11\linewidth}\centering
            \includegraphics[width=\linewidth,height=\linewidth]{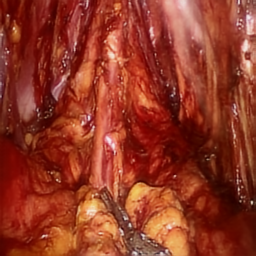}
        \end{minipage} &
        \begin{minipage}[b]{0.11\linewidth}\centering
            \includegraphics[width=\linewidth,height=\linewidth]{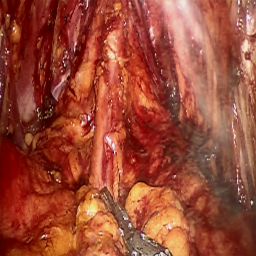}
        \end{minipage} &
        \begin{minipage}[b]{0.11\linewidth}\centering
            \includegraphics[width=\linewidth,height=\linewidth]{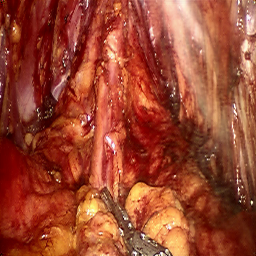}
        \end{minipage} &
        \begin{minipage}[b]{0.11\linewidth}\centering
            \includegraphics[width=\linewidth,height=\linewidth]{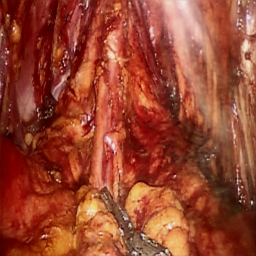}
        \end{minipage} &
        \begin{minipage}[b]{0.11\linewidth}\centering
            \includegraphics[width=\linewidth,height=\linewidth]{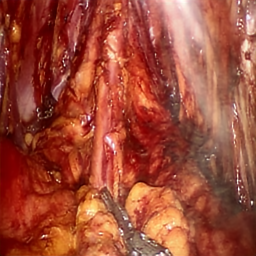} 
        \end{minipage} &
        \begin{minipage}[b]{0.11\linewidth}\centering
            \includegraphics[width=\linewidth,height=\linewidth]{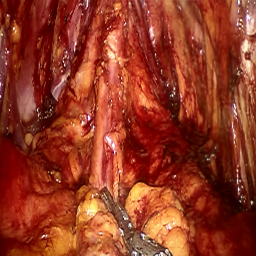}
        \end{minipage} &
        \begin{minipage}[b]{0.11\linewidth}\centering
            \includegraphics[width=\linewidth,height=\linewidth]{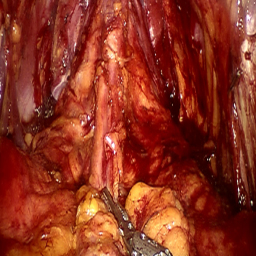}
        \end{minipage} \\

           \begin{minipage}[b]{0.11\linewidth}\centering
            \includegraphics[width=\linewidth,height=\linewidth]{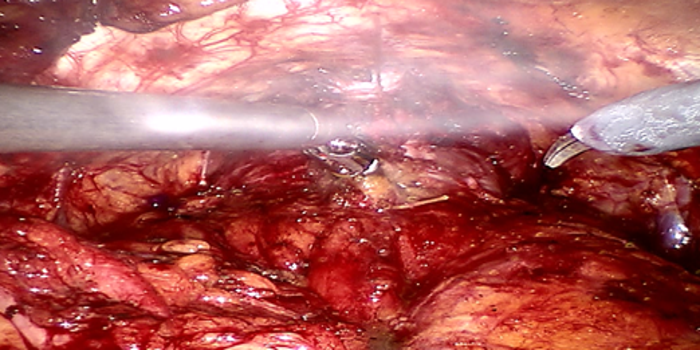}
        \end{minipage} &
        \begin{minipage}[b]{0.11\linewidth}\centering
            \includegraphics[width=\linewidth,height=\linewidth]{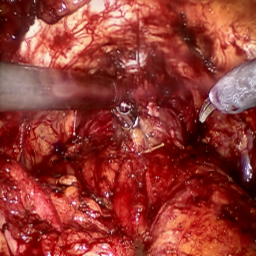} 
        \end{minipage} &
        \begin{minipage}[b]{0.11\linewidth}\centering
            \includegraphics[width=\linewidth,height=\linewidth]{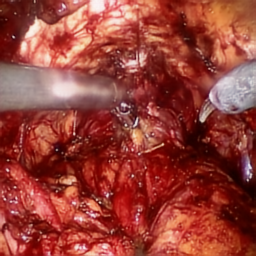}
        \end{minipage} &
        \begin{minipage}[b]{0.11\linewidth}\centering
            \includegraphics[width=\linewidth,height=\linewidth]{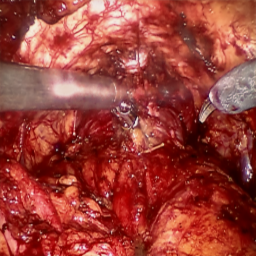}
        \end{minipage} &
        \begin{minipage}[b]{0.11\linewidth}\centering
            \includegraphics[width=\linewidth,height=\linewidth]{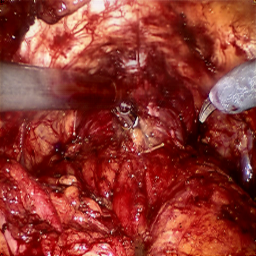}
        \end{minipage} &
        \begin{minipage}[b]{0.11\linewidth}\centering
            \includegraphics[width=\linewidth,height=\linewidth]{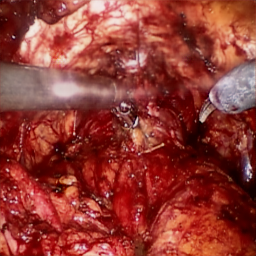}
        \end{minipage} &
        \begin{minipage}[b]{0.11\linewidth}\centering
            \includegraphics[width=\linewidth,height=\linewidth]{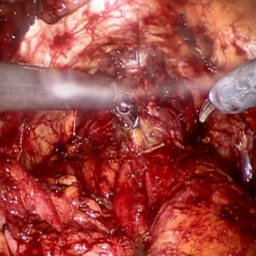} 
        \end{minipage} &
        \begin{minipage}[b]{0.11\linewidth}\centering
            \includegraphics[width=\linewidth,height=\linewidth]{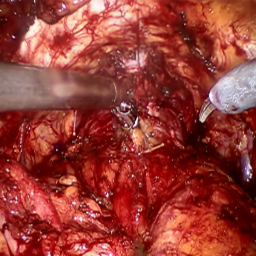}
        \end{minipage} &
        \begin{minipage}[b]{0.11\linewidth}\centering
            \includegraphics[width=\linewidth,height=\linewidth]{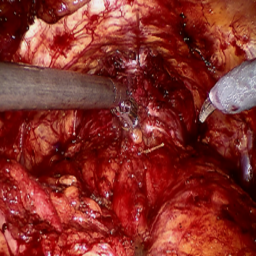}

        \end{minipage} \\

        \begin{minipage}[b]{0.11\linewidth}\centering
            \includegraphics[width=\linewidth,height=\linewidth]{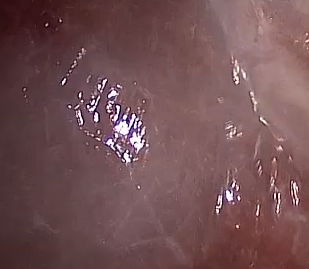}
        \end{minipage} &
        \begin{minipage}[b]{0.11\linewidth}\centering
            \includegraphics[width=\linewidth,height=\linewidth]{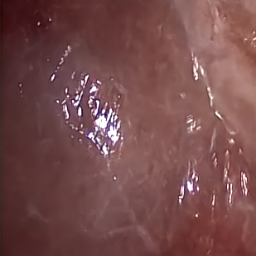} 
        \end{minipage} &
        \begin{minipage}[b]{0.11\linewidth}\centering
            \includegraphics[width=\linewidth,height=\linewidth]{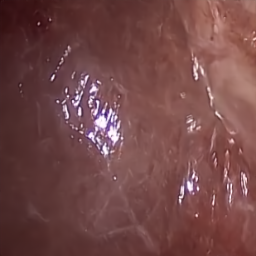}
        \end{minipage} &
        \begin{minipage}[b]{0.11\linewidth}\centering
            \includegraphics[width=\linewidth,height=\linewidth]{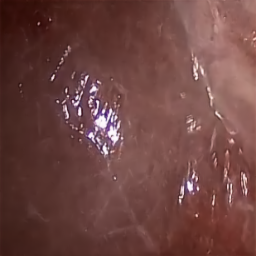}
        \end{minipage} &
        \begin{minipage}[b]{0.11\linewidth}\centering
            \includegraphics[width=\linewidth,height=\linewidth]{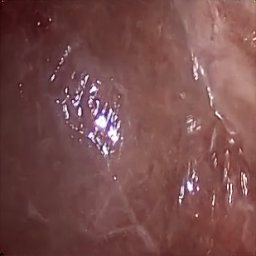}
        \end{minipage} &
        \begin{minipage}[b]{0.11\linewidth}\centering
            \includegraphics[width=\linewidth,height=\linewidth]{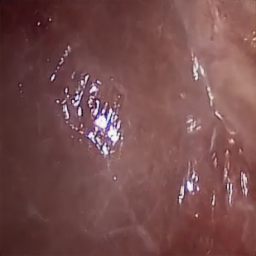}
        \end{minipage} &
        \begin{minipage}[b]{0.11\linewidth}\centering
            \includegraphics[width=\linewidth,height=\linewidth]{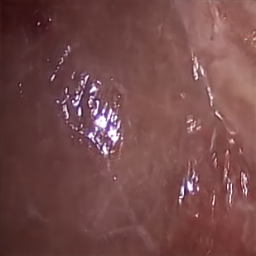} 
        \end{minipage} &
        \begin{minipage}[b]{0.11\linewidth}\centering
            \includegraphics[width=\linewidth,height=\linewidth]{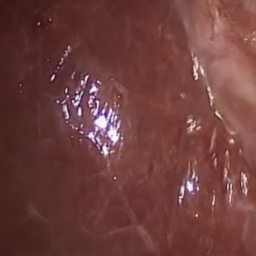}
        \end{minipage} &
        \begin{minipage}[b]{0.11\linewidth}\centering
            \includegraphics[width=\linewidth,height=\linewidth]{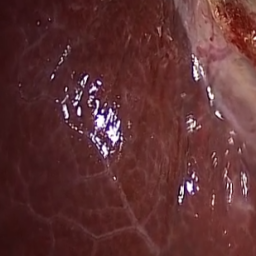}
        \end{minipage} \\
        
  \begin{minipage}[b]{0.11\linewidth}\centering
    \includegraphics[width=\linewidth,height=\linewidth]{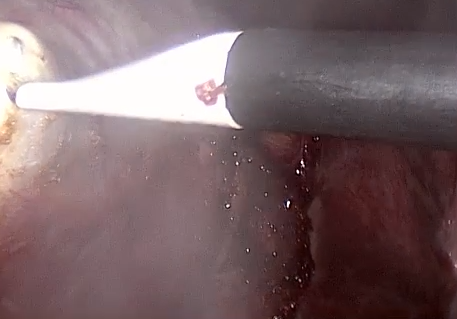}
    \centerline{\scriptsize (a) Input}
\end{minipage} &
\begin{minipage}[b]{0.11\linewidth}\centering
    \includegraphics[width=\linewidth,height=\linewidth]{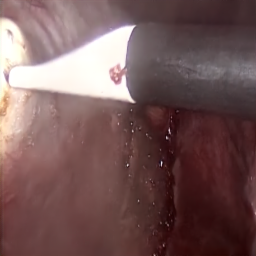}
    \centerline{\scriptsize (b) DIACMPN}
\end{minipage} &
\begin{minipage}[b]{0.11\linewidth}\centering
    \includegraphics[width=\linewidth,height=\linewidth]{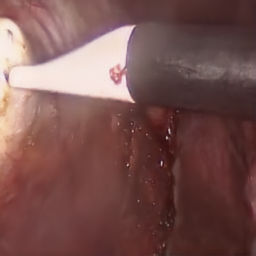}
    \centerline{\scriptsize (c) Desmoke-LAP}
\end{minipage} &
\begin{minipage}[b]{0.11\linewidth}\centering
    \includegraphics[width=\linewidth,height=\linewidth]{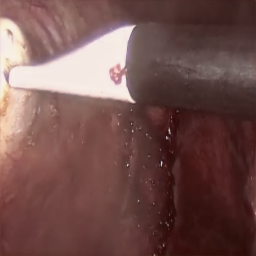}
    \centerline{\scriptsize (d) PFAN}
\end{minipage} &
\begin{minipage}[b]{0.11\linewidth}\centering
    \includegraphics[width=\linewidth,height=\linewidth]{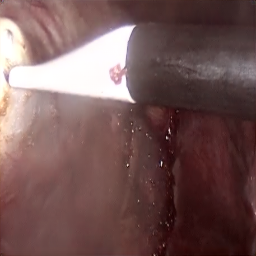}
    \centerline{\scriptsize (e) MITNET}
\end{minipage} &
\begin{minipage}[b]{0.11\linewidth}\centering
    \includegraphics[width=\linewidth,height=\linewidth]{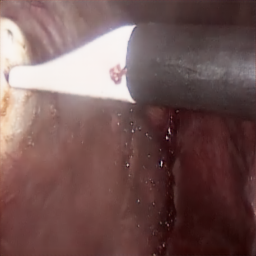}
    \centerline{\scriptsize (f) Salazar}
\end{minipage} &
\begin{minipage}[b]{0.11\linewidth}\centering
    \includegraphics[width=\linewidth,height=\linewidth]{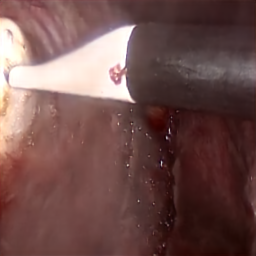}
    \centerline{\scriptsize (g) Dehamer}
\end{minipage} &
\begin{minipage}[b]{0.11\linewidth}\centering
    \includegraphics[width=\linewidth,height=\linewidth]{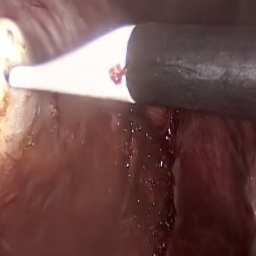}
    \centerline{\scriptsize (h) Ours}
\end{minipage} &
\begin{minipage}[b]{0.11\linewidth}\centering
    \includegraphics[width=\linewidth,height=\linewidth]{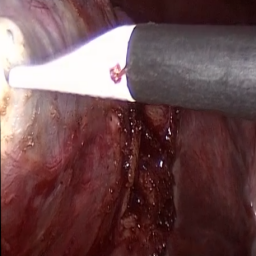}
    \centerline{\scriptsize (i) Target}
\end{minipage} \\

    \end{tabular}
    \caption{Comparison of different methods on the DesmokeData dataset.}
    \label{fig:comp2}
\end{figure*}
\begin{figure*}[htbp]
    \centering
    \setlength{\tabcolsep}{1pt} 
    \renewcommand{\arraystretch}{0.9} 
    \begin{tabular}{ccccccccc}
        \begin{minipage}[b]{0.11\linewidth}\centering
            \includegraphics[width=\linewidth,height=\linewidth]{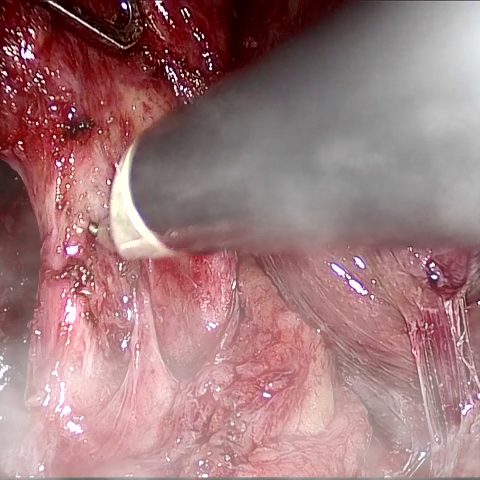}
        \end{minipage} &
        \begin{minipage}[b]{0.11\linewidth}\centering
            \includegraphics[width=\linewidth,height=\linewidth]{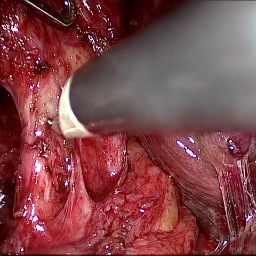} 
        \end{minipage} &
        \begin{minipage}[b]{0.11\linewidth}\centering
            \includegraphics[width=\linewidth,height=\linewidth]{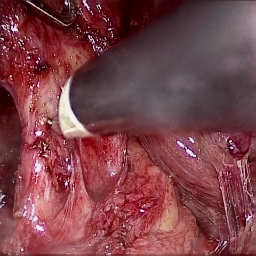}
        \end{minipage} &
        \begin{minipage}[b]{0.11\linewidth}\centering
            \includegraphics[width=\linewidth,height=\linewidth]{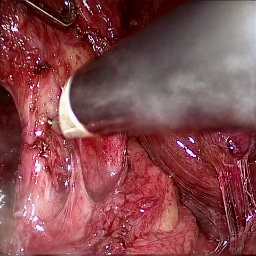}
        \end{minipage} &
        \begin{minipage}[b]{0.11\linewidth}\centering
            \includegraphics[width=\linewidth,height=\linewidth]{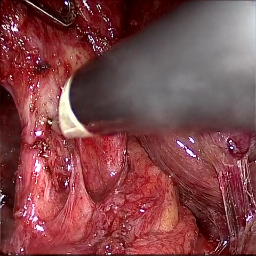}
        \end{minipage} &
        \begin{minipage}[b]{0.11\linewidth}\centering
            \includegraphics[width=\linewidth,height=\linewidth]{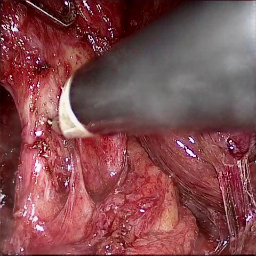}
        \end{minipage} &
        \begin{minipage}[b]{0.11\linewidth}\centering
            \includegraphics[width=\linewidth,height=\linewidth]{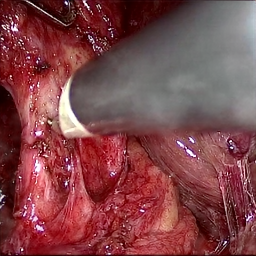} 
        \end{minipage} &
        \begin{minipage}[b]{0.11\linewidth}\centering
            \includegraphics[width=\linewidth,height=\linewidth]{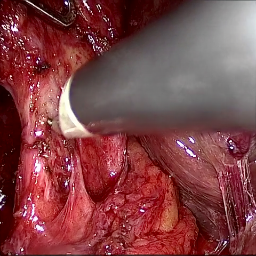}
        \end{minipage} &
        \begin{minipage}[b]{0.11\linewidth}\centering
            \includegraphics[width=\linewidth,height=\linewidth]{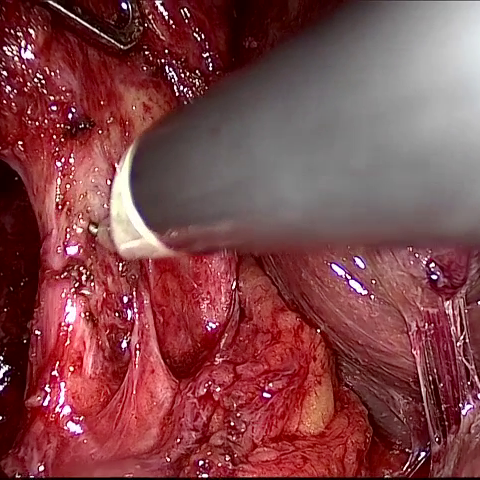}
        \end{minipage} \\

        \begin{minipage}[b]{0.11\linewidth}\centering
            \includegraphics[width=\linewidth,height=\linewidth]{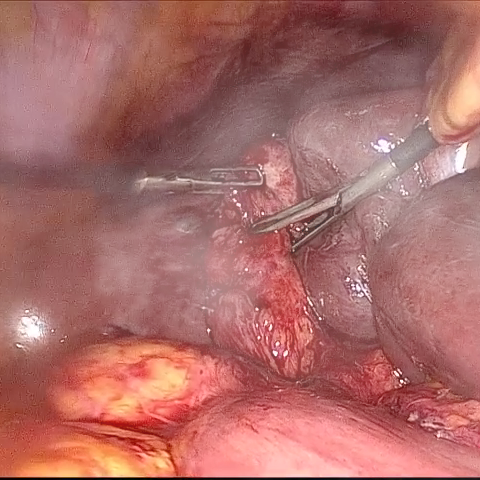}
        \end{minipage} &
        \begin{minipage}[b]{0.11\linewidth}\centering
            \includegraphics[width=\linewidth,height=\linewidth]{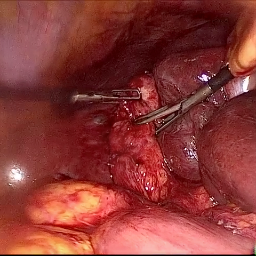}
        \end{minipage} &
        \begin{minipage}[b]{0.11\linewidth}\centering
            \includegraphics[width=\linewidth,height=\linewidth]{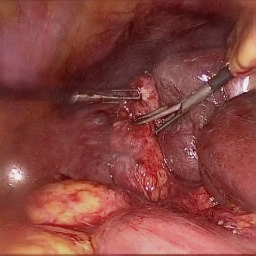}
        \end{minipage} &
        \begin{minipage}[b]{0.11\linewidth}\centering
            \includegraphics[width=\linewidth,height=\linewidth]{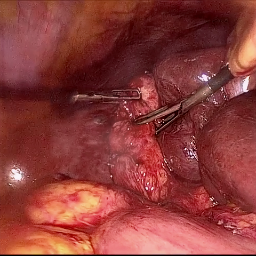}
        \end{minipage} &
        \begin{minipage}[b]{0.11\linewidth}\centering
            \includegraphics[width=\linewidth,height=\linewidth]{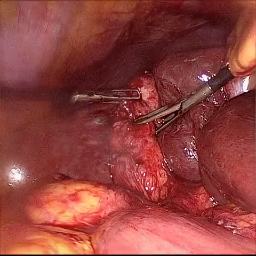}
        \end{minipage} &
        \begin{minipage}[b]{0.11\linewidth}\centering
            \includegraphics[width=\linewidth,height=\linewidth]{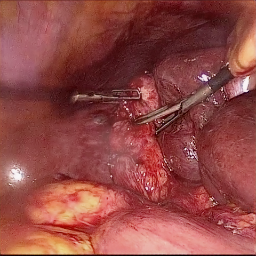}
        \end{minipage} &
        \begin{minipage}[b]{0.11\linewidth}\centering
            \includegraphics[width=\linewidth,height=\linewidth]{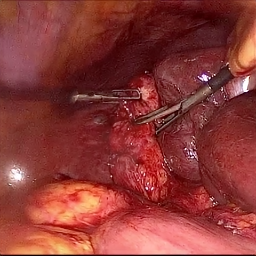}
        \end{minipage} &
        \begin{minipage}[b]{0.11\linewidth}\centering
            \includegraphics[width=\linewidth,height=\linewidth]{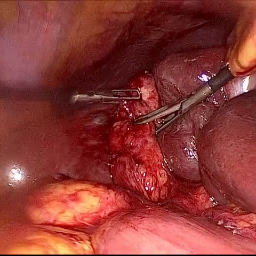}
        \end{minipage} &
        \begin{minipage}[b]{0.11\linewidth}\centering
            \includegraphics[width=\linewidth,height=\linewidth]{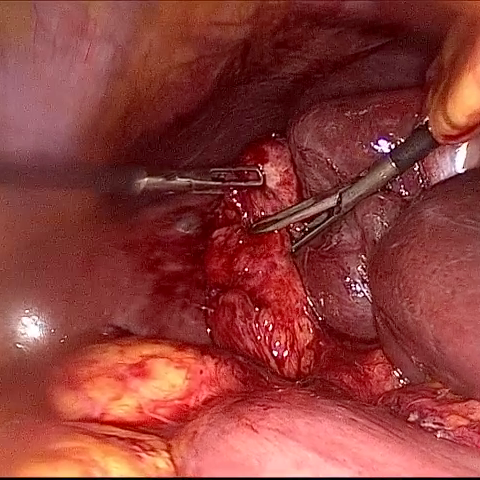}
        \end{minipage} \\

        \begin{minipage}[b]{0.11\linewidth}\centering
            \includegraphics[width=\linewidth,height=\linewidth]{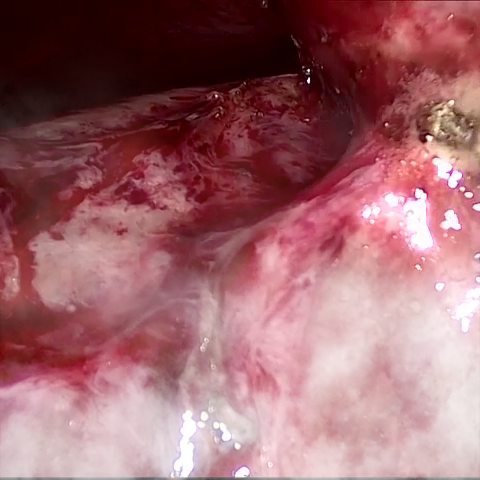}
        \end{minipage} &
        \begin{minipage}[b]{0.11\linewidth}\centering
            \includegraphics[width=\linewidth,height=\linewidth]{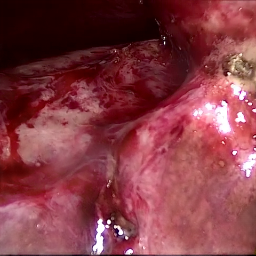}
        \end{minipage} &
        \begin{minipage}[b]{0.11\linewidth}\centering
            \includegraphics[width=\linewidth,height=\linewidth]{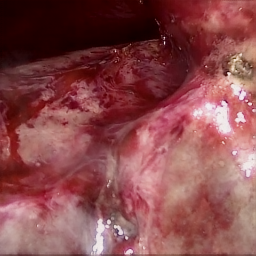}
        \end{minipage} &
        \begin{minipage}[b]{0.11\linewidth}\centering
            \includegraphics[width=\linewidth,height=\linewidth]{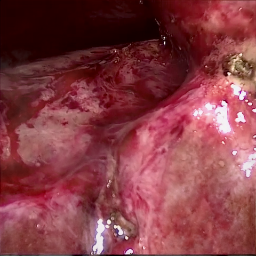}
        \end{minipage} &
        \begin{minipage}[b]{0.11\linewidth}\centering
            \includegraphics[width=\linewidth,height=\linewidth]{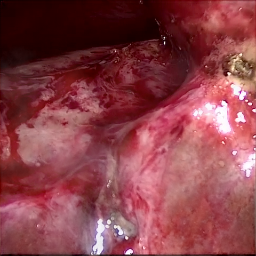}
        \end{minipage} &
        \begin{minipage}[b]{0.11\linewidth}\centering
            \includegraphics[width=\linewidth,height=\linewidth]{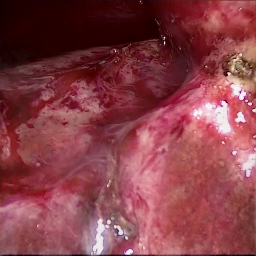}
        \end{minipage} &
        \begin{minipage}[b]{0.11\linewidth}\centering
            \includegraphics[width=\linewidth,height=\linewidth]{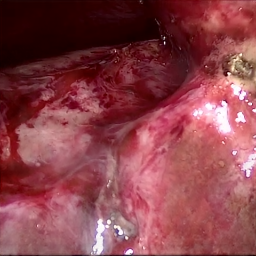}
        \end{minipage} &
        \begin{minipage}[b]{0.11\linewidth}\centering
            \includegraphics[width=\linewidth,height=\linewidth]{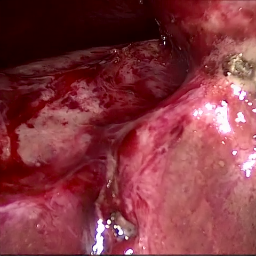}
        \end{minipage} &
        \begin{minipage}[b]{0.11\linewidth}\centering
            \includegraphics[width=\linewidth,height=\linewidth]{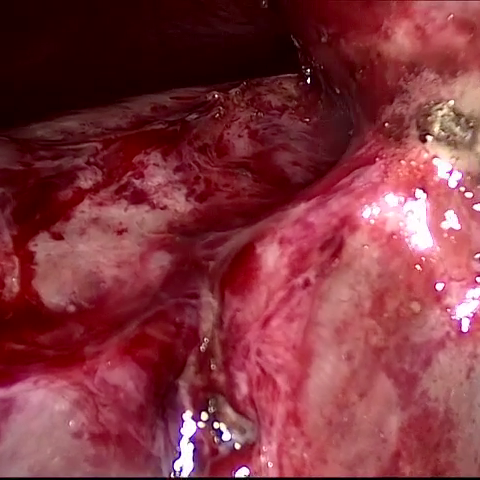}
        \end{minipage} \\

\begin{minipage}[b]{0.11\linewidth}\centering
    \includegraphics[width=\linewidth,height=\linewidth]{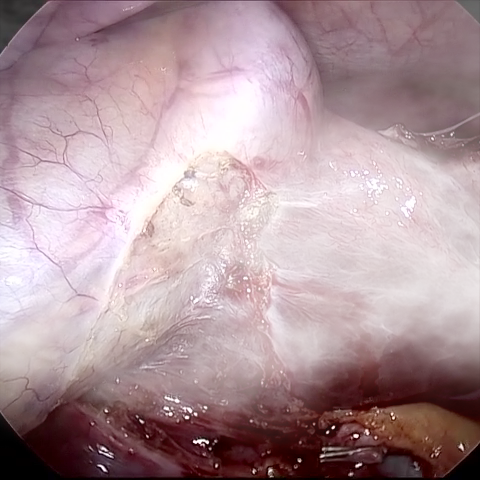}
    \centerline{\scriptsize (a) Input}
\end{minipage} &
\begin{minipage}[b]{0.11\linewidth}\centering
    \includegraphics[width=\linewidth,height=\linewidth]{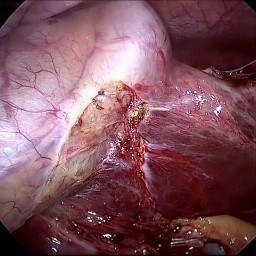}
    \centerline{\scriptsize (b) DIACMPN}
\end{minipage} &
\begin{minipage}[b]{0.11\linewidth}\centering
    \includegraphics[width=\linewidth,height=\linewidth]{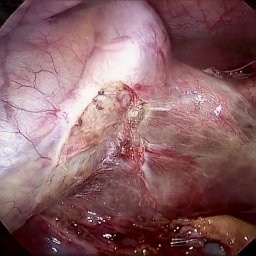}
    \centerline{\scriptsize (c) Desmoke-LAP}
\end{minipage} &
\begin{minipage}[b]{0.11\linewidth}\centering
    \includegraphics[width=\linewidth,height=\linewidth]{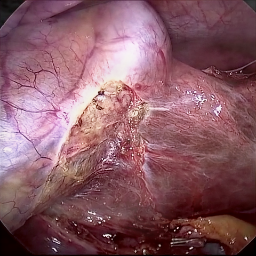}
    \centerline{\scriptsize (d) PFAN}
\end{minipage} &
\begin{minipage}[b]{0.11\linewidth}\centering
    \includegraphics[width=\linewidth,height=\linewidth]{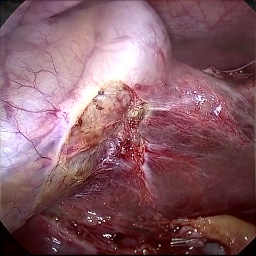}
    \centerline{\scriptsize (e) MITNET}
\end{minipage} &
\begin{minipage}[b]{0.11\linewidth}\centering
    \includegraphics[width=\linewidth,height=\linewidth]{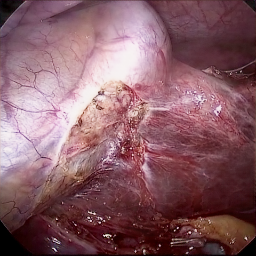}
    \centerline{\scriptsize (f) Salazar}
\end{minipage} &
\begin{minipage}[b]{0.11\linewidth}\centering
    \includegraphics[width=\linewidth,height=\linewidth]{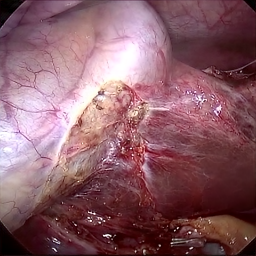}
    \centerline{\scriptsize (g) Dehamer}
\end{minipage} &
\begin{minipage}[b]{0.11\linewidth}\centering
    \includegraphics[width=\linewidth,height=\linewidth]{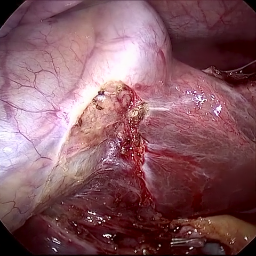}
    \centerline{\scriptsize (h) Ours}
\end{minipage} &
\begin{minipage}[b]{0.11\linewidth}\centering
    \includegraphics[width=\linewidth,height=\linewidth]{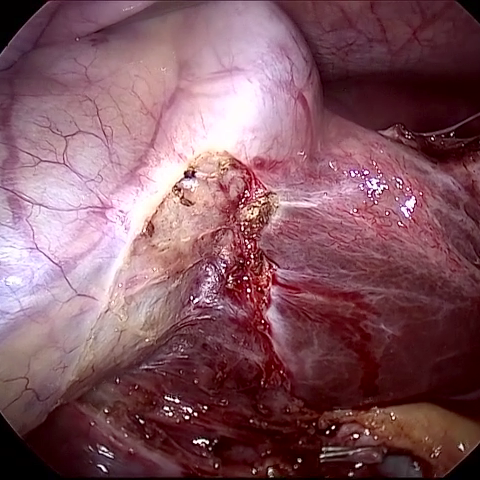}
    \centerline{\scriptsize (i) Target}
\end{minipage} \\

    \end{tabular}
    \caption{Comparison of different methods on the LSD3K dataset.}
    \label{fig:comp1}
\end{figure*}

\subsection{Dataset and Evaluation Metrics}
For our experiments, we utilize two publicly available datasets. The first is DesmokeData~\cite{xia2024new}, a collection of \textit{in vivo} laparoscopic images. This dataset is divided into a training set of 2,952 paired images and a testing set of 512 paired images. The second dataset is LSD3K~\cite{chang2024lsd3k}, which is a larger benchmark for smoke removal in surgical images. We follow the standard split, using 2,800 images for training and 200 for testing.

To evaluate the performance of our model, we use two widely adopted image quality assessment metrics: the Peak Signal-to-Noise Ratio (PSNR) and the Structural Similarity Index Measure (SSIM)~\cite{wang2004image}. PSNR measures the pixel-wise reconstruction accuracy, while SSIM assesses the perceptual similarity between the ground truth and the restored images.

\subsection{Implementation Details}
We implement our model in PyTorch and train it on 2 NVIDIA A800 80G GPUs. The training process uses $256\times256$ input images with a batch size of 4, employing the AdamW optimizer~\cite{loshchilov2018decoupled} (initial learning rate $2\times10^{-4}$) and the Cosine Annealing scheduler~\cite{loshchilov2017sgdr}. We choose the AdamW optimizer for its decoupled weight decay to mitigate overfitting in transformer-based attention modules~\cite{cxh2,cxh4}, and the Cosine Annealing scheduler to avoid local minima and boost generalization across surgical smoke datasets. To enhance model robustness, we apply standard data augmentation techniques, including random cropping, flipping, rotation, and mix-up. For testing, images are resized to $256\times256$ resolution.

\begin{table*}[ht]
\centering
\caption{Quantitative comparison with state-of-the-art methods on the LSD3K and DesmokeData datasets. The best results are highlighted in \textbf{bold}.}
\adjustbox{max width=\linewidth}{%
\begin{tabular}{l|cccccc|cccccc}
\toprule
\multirow{2}{*}{Method} & \multicolumn{6}{c|}{DesmokeData}                           & \multicolumn{6}{c}{LSD3K}                                  \\ \cmidrule(lr){2-13}
                        & PSNR\textuparrow    & SSIM\textuparrow  & MS-SSIM\textuparrow & MAE\textdownarrow    & LPIPS\textdownarrow & CIEDE-2000\textdownarrow & PSNR\textuparrow    & SSIM\textuparrow  & MS-SSIM\textuparrow & MAE\textdownarrow    & LPIPS\textdownarrow & CIEDE-2000\textdownarrow  \\ \midrule
AECRNet~\cite{wu2021contrastive}                 & 26.446 & 0.920 & 0.954  & 9.352  & 0.074 & 0.052     & 24.925  & 0.840   & 0.917  & 11.601 & 0.109 & 0.067     \\
AODNet~\cite{li2017aod}                  & 16.186  & 0.769 & 0.807  & 31.530   & 0.200 & 0.150       & 21.356 & 0.786 & 0.871  & 20.276 & 0.146  & 0.102     \\
Salazar~\etal~\cite{salazar2020desmoking}                 & 25.052 & 0.890   & 0.940  & 11.394 & 0.086 & 0.066     & 24.599 & 0.829 & 0.905  & 12.904 & 0.116 & 0.072     \\
DCP~\cite{he2010single}                     & 16.620 & 0.775 & 0.802   & 30.345 & 0.168 & 0.151     & 20.882 & 0.779 & 0.864  & 20.233 & 0.153 & 0.115     \\
Dehamer~\cite{guo2022image}                 & 27.603 & 0.918 & 0.955  & 8.261  & 0.073 & 0.044     & 23.629 & 0.815 & 0.894  & 14.595 & 0.129 & 0.078     \\
DehazeFormer~\cite{song2023vision}            & 29.416 & 0.938 & 0.971  & 6.236  & 0.060 & 0.031     & 25.682 & 0.849 & 0.923  & 10.864 & 0.104 & 0.057     \\
DehazeNet~\cite{cai2016dehazenet}               & 17.376 & 0.774 & 0.797  & 26.601 & 0.186 & 0.162     & 21.525 & 0.783 & 0.876  & 19.281 & 0.150 & 0.106     \\
Desmoke-CycleGAN~\cite{hu2021cycle} & 20.847 & 0.834 & 0.903 & 17.855 & 0.114 & 0.107 & 24.220 & 0.808 & 0.911 & 12.502 & 0.153 & 0.077 \\
Desmoke-LAP~\cite{pan2022desmoke}              & 23.546 & 0.888 & 0.931  & 13.191 & 0.094 & 0.081     & 24.515 & 0.828 & 0.916  & 12.228 & 0.117 & 0.073     \\
DesmokeNet~\cite{chen2021desmokenet}              & 26.271 & 0.903 & 0.949  & 9.522  & 0.079 & 0.056     & 24.972 & 0.835 & 0.912  & 12.067 & 0.118 & 0.070     \\
DIACMPN~\cite{zhang2024depth}                 & 28.332 & 0.929 & 0.965  & 7.314  & 0.064 & 0.036     & 25.557 & 0.845 & 0.920  & 11.045 & 0.107 & 0.061     \\
FFANet~\cite{qin2020ffa}                  & 29.354 & 0.935  & 0.971  & 6.114  & 0.060 & 0.031     & 25.613 & 0.851 & 0.920  & 10.967 & 0.105 & 0.059     \\
GridDehazeNet~\cite{liu2019griddehazenet}           & 16.688  & 0.773 & 0.775  & 29.315 & 0.214 & 0.157     & 24.370 & 0.837 & 0.906  & 12.949  & 0.117 & 0.069     \\
Bolkar~\etal~\cite{bolkar2018deep}                    & 7.015  & 0.589 & 0.600  & 79.021 & 0.828  & 1.391     & 6.724  & 0.569 & 0.652   & 77.771 & 0.870 & 1.460     \\
MITNet~\cite{shen2023mutual}                  & 26.725 & 0.904 & 0.949  & 9.446  & 0.077 & 0.049      & 25.384 & 0.837 & 0.916  & 11.306 & 0.105 & 0.062     \\
MSBDN~\cite{dong2020multi}                   & 27.187 & 0.916 & 0.957  & 8.647  & 0.072 & 0.047      & 25.560 & 0.845 & 0.918  & 11.147 & 0.110 & 0.064     \\
PFAN~\cite{zhang2023progressive}                    & 25.041 & 0.895 & 0.938  & 11.560 & 0.086  & 0.062     & 24.184  & 0.827 & 0.902  & 13.352 & 0.115 & 0.073     \\
SFSNiD~\cite{cong2024semi}                  & 28.600 & 0.933 & 0.966  & 6.917  & 0.061 & 0.034     & 25.666 & 0.851 & 0.922  & 10.672 & 0.104 & 0.060       \\
Sidorov~\etal~\cite{sidorov2020generative}                 & 24.857 & 0.890   & 0.939  & 11.587  & 0.085 & 0.066     & 24.544 & 0.825 & 0.905  & 12.971 & 0.118 & 0.072     \\ \midrule
Ours                   & \textbf{30.258} & \textbf{0.945} & \textbf{0.977}  & \textbf{5.445}  & \textbf{0.055} & \textbf{0.028}     & \textbf{25.820} & \textbf{0.855}  & \textbf{0.927}  & \textbf{10.516} & \textbf{0.101} & \textbf{0.058}     \\
\bottomrule
\end{tabular}%
}
\label{tab:comp}
\end{table*}

\subsection{Comparisons with State-of-the-arts}
To validate the superiority of our proposed RGA-Net, we conduct a comprehensive comparison against a range of state-of-the-art (SOTA) image dehazing and surgical smoke removal methods. The selected methods include the classic prior-based method DCP~\cite{he2010single}; prominent CNN-based architectures such as AOD-Net~\cite{li2017aod}, FFA-Net~\cite{qin2020ffa}, and MSBDN~\cite{dong2020multi}; and recent powerful Transformer-based models like DehazeFormer~\cite{song2023vision} and the surgical-specific PFAN~\cite{zhang2023progressive}. We evaluate all methods on the DesmokeData~\cite{xia2024new} and LSD3K~\cite{chang2024lsd3k} test sets using PSNR and SSIM as the evaluation metrics.

\subsubsection{Quantitative Analysis}

The quantitative results are summarized in \cref{tab:comp}. On both datasets, our RGA-Net consistently outperforms all competing methods across all metrics. On the DesmokeData dataset, RGA-Net achieves a PSNR of 30.258 dB and an SSIM of 0.945 , marking a significant improvement of 0.842 dB in PSNR and 0.007 in SSIM over the next-best method, DehazeFormer (which achieves 29.416 dB and 0.938).

Similarly, on the LSD3K dataset, RGA-Net establishes a new state-of-the-art with a PSNR of 25.820 dB and an SSIM of 0.855. While Transformer-based models like SFSNiD and DehazeFormer perform strongly, our specialized architecture proves to be more effective for this specific domain. The superior performance confirms that combining hybrid attention mechanisms and reciprocal gating handles the complex nature of surgical smoke more robustly than general-purpose dehazing algorithms.

\subsubsection{Qualitative Analysis}
\Cref{fig:comp1,fig:comp2} provide a visual comparison of the desmoking results on challenging images from the test sets. The visual results corroborate our quantitative findings. 
DCP not only fails to remove the dense smoke but also introduces severe color distortion. CNN-based methods like FFA-Net and MSBDN successfully remove a large portion of the smoke but tend to either leave behind a thin layer of residual haze or over-smooth the image, losing critical textural details of the tissue and surgical instruments. The Transformer-based methods, DehazeFormer and PFAN, produce significantly better results by restoring more details. However, they can sometimes struggle with non-uniform smoke distribution, resulting in regions with unnatural brightness or minor artifacts.

In stark contrast, our RGA-Net generates visually superior results that are remarkably close to the ground-truth images. It effectively removes even the densest plumes of smoke while simultaneously preserving fine-grained details, such as blood vessels, tissue textures, and reflections on surgical tools. Furthermore, the color and illumination of the restored scene appear more natural and consistent, which is a direct benefit of our model's ability to handle both local and global features through its hybrid attention and cross-gating mechanisms. This enhanced visual clarity is crucial for improving the surgeon-robot interface in real-world clinical applications.

\begin{table*}[ht]
\centering
\caption{Ablation studies on the LSD3K and DesmokeData datasets. The best results are highlighted in \textbf{bold}.}
\adjustbox{max width=\linewidth}{%
\begin{tabular}{ccc|cccccc|cccccc}
\toprule
\multicolumn{3}{c|}{Module} & \multicolumn{6}{c|}{DesmokeData}                 & \multicolumn{6}{c}{LSD3K}                       \\ \cmidrule(lr){1-3}
DHA & ADA & CrossGating & PSNR & SSIM & MS-SSIM & MAE & LPIPS & CIEDE-2000 & PSNR & SSIM & MS-SSIM & MAE & LPIPS & CIEDE-2000 \\ \midrule
       \XSolidBrush &  \Checkmark  &  \Checkmark  & 23.796 & 0.824 & 0.897 & 14.297 & 0.114 & 5.337  & 23.225 & 0.876 & 0.924 & 14.185 & 0.090 & 5.190 \\
       \Checkmark & \XSolidBrush  &  \Checkmark       & 22.430 & 0.802 & 0.884 & 17.498 & 0.128 & 6.451  & 18.921 & 0.811 & 0.862 & 23.071 & 0.153 & 8.892 \\
       \Checkmark &  \Checkmark       &  \XSolidBrush       & 8.471  & 0.175 & 0.198 & 76.234 & 0.895 & 27.116 & 19.273 & 0.820 & 0.857 & 22.323 & 0.153 & 9.265 \\
      \Checkmark  &  \Checkmark       &  \Checkmark       & \textbf{30.258} & \textbf{0.945} & \textbf{0.977}  & \textbf{5.445}  & \textbf{0.055} & \textbf{0.028}     & \textbf{25.820} & \textbf{0.855}  & \textbf{0.927}  & \textbf{10.516} & \textbf{0.101} & \textbf{0.058}  \\ \bottomrule
\end{tabular}%
}
\label{tab:ab}
\end{table*}

\begin{figure*}[htbp]
    \centering
    \setlength{\tabcolsep}{1pt} 
    \renewcommand{\arraystretch}{0.9} 
    \begin{tabular}{ccccccccc}

        \begin{minipage}[b]{0.16\linewidth}\centering
            \includegraphics[width=\linewidth,height=\linewidth]{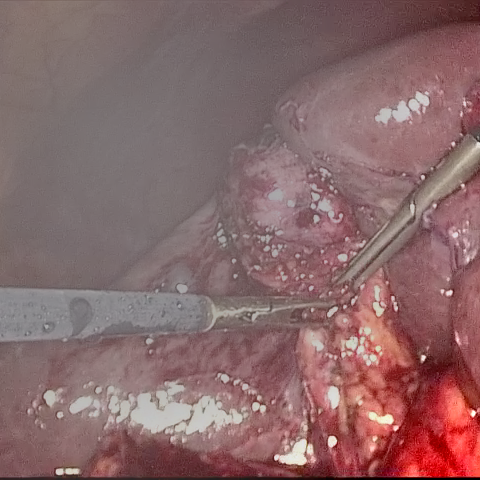}
        \end{minipage} &
        \begin{minipage}[b]{0.16\linewidth}\centering
            \includegraphics[width=\linewidth,height=\linewidth]{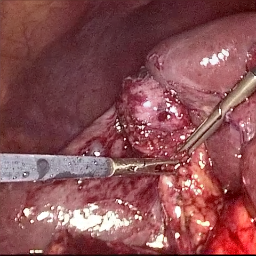} 
        \end{minipage} &
        \begin{minipage}[b]{0.16\linewidth}\centering
            \includegraphics[width=\linewidth,height=\linewidth]{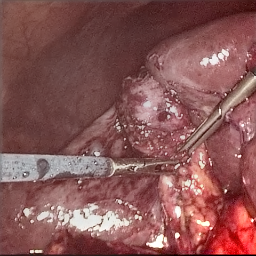} 
        \end{minipage} &
        \begin{minipage}[b]{0.16\linewidth}\centering
            \includegraphics[width=\linewidth,height=\linewidth]{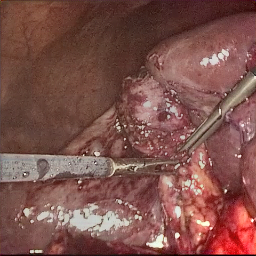} 
        \end{minipage} &
        \begin{minipage}[b]{0.16\linewidth}\centering
            \includegraphics[width=\linewidth,height=\linewidth]{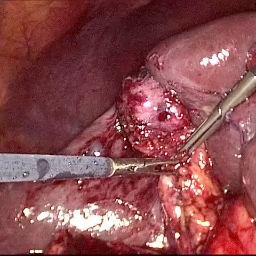}
        \end{minipage} &
        \begin{minipage}[b]{0.16\linewidth}\centering
            \includegraphics[width=\linewidth,height=\linewidth]{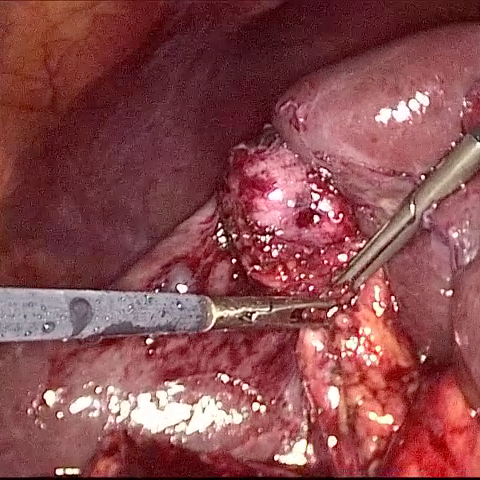}
        \end{minipage}  \\

        \begin{minipage}[b]{0.16\linewidth}\centering
            \includegraphics[width=\linewidth,height=\linewidth]{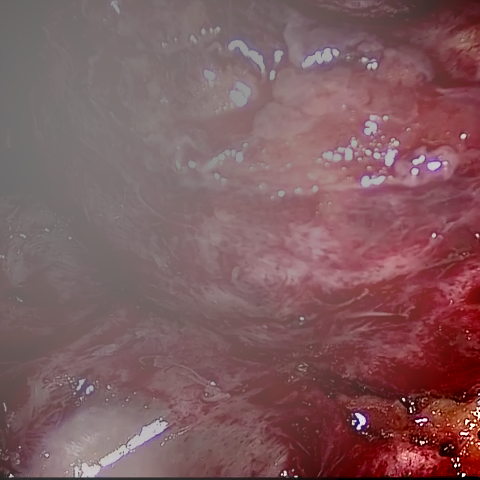}
        \end{minipage} &
        \begin{minipage}[b]{0.16\linewidth}\centering
            \includegraphics[width=\linewidth,height=\linewidth]{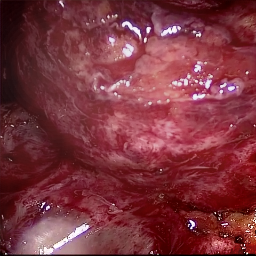} 
        \end{minipage} &
        \begin{minipage}[b]{0.16\linewidth}\centering
            \includegraphics[width=\linewidth,height=\linewidth]{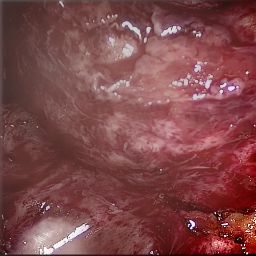} 
        \end{minipage} &
        \begin{minipage}[b]{0.16\linewidth}\centering
            \includegraphics[width=\linewidth,height=\linewidth]{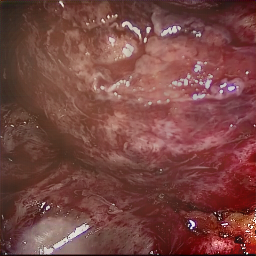} 
        \end{minipage} &
        \begin{minipage}[b]{0.16\linewidth}\centering
            \includegraphics[width=\linewidth,height=\linewidth]{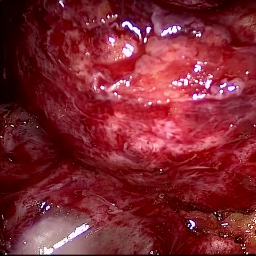}
        \end{minipage} &
        \begin{minipage}[b]{0.16\linewidth}\centering
            \includegraphics[width=\linewidth,height=\linewidth]{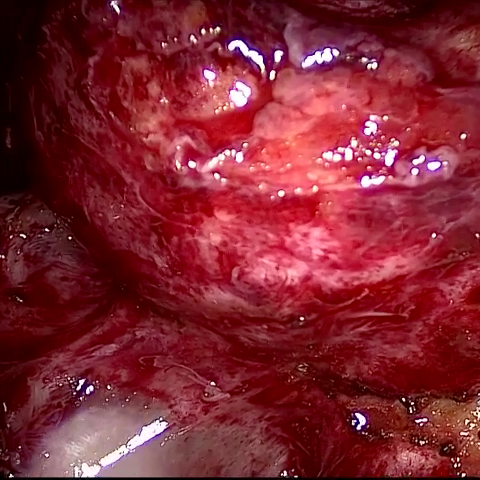}
        \end{minipage}  \\

        \begin{minipage}[b]{0.16\linewidth}\centering
            \includegraphics[width=\linewidth,height=\linewidth]{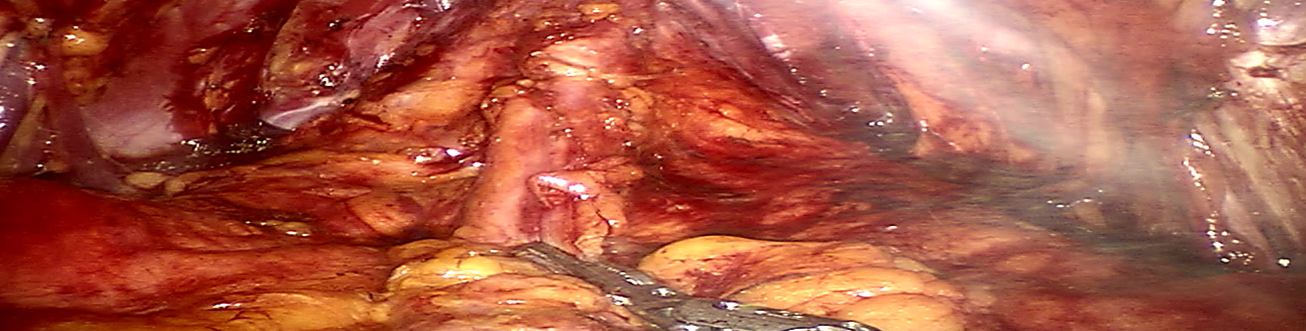}
        \end{minipage} &
        \begin{minipage}[b]{0.16\linewidth}\centering
            \includegraphics[width=\linewidth,height=\linewidth]{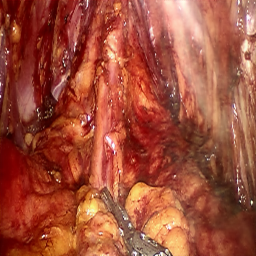} 
        \end{minipage} &
        \begin{minipage}[b]{0.16\linewidth}\centering
            \includegraphics[width=\linewidth,height=\linewidth]{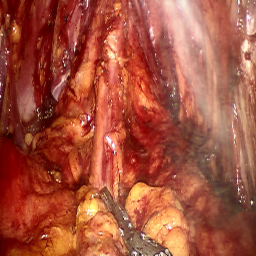} 
        \end{minipage} &
        \begin{minipage}[b]{0.16\linewidth}\centering
            \includegraphics[width=\linewidth,height=\linewidth]{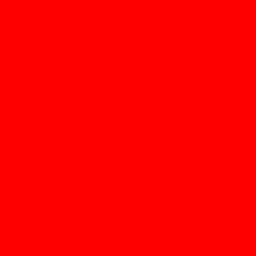} 
        \end{minipage} &
        \begin{minipage}[b]{0.16\linewidth}\centering
            \includegraphics[width=\linewidth,height=\linewidth]{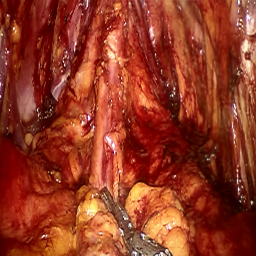}
        \end{minipage} &
        \begin{minipage}[b]{0.16\linewidth}\centering
            \includegraphics[width=\linewidth,height=\linewidth]{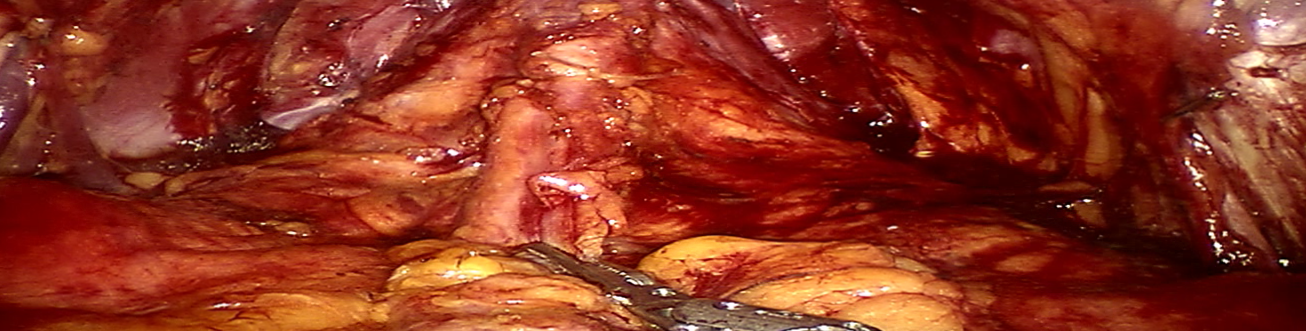}
        \end{minipage}  \\

        \begin{minipage}[b]{0.16\linewidth}\centering
            \includegraphics[width=\linewidth,height=\linewidth]{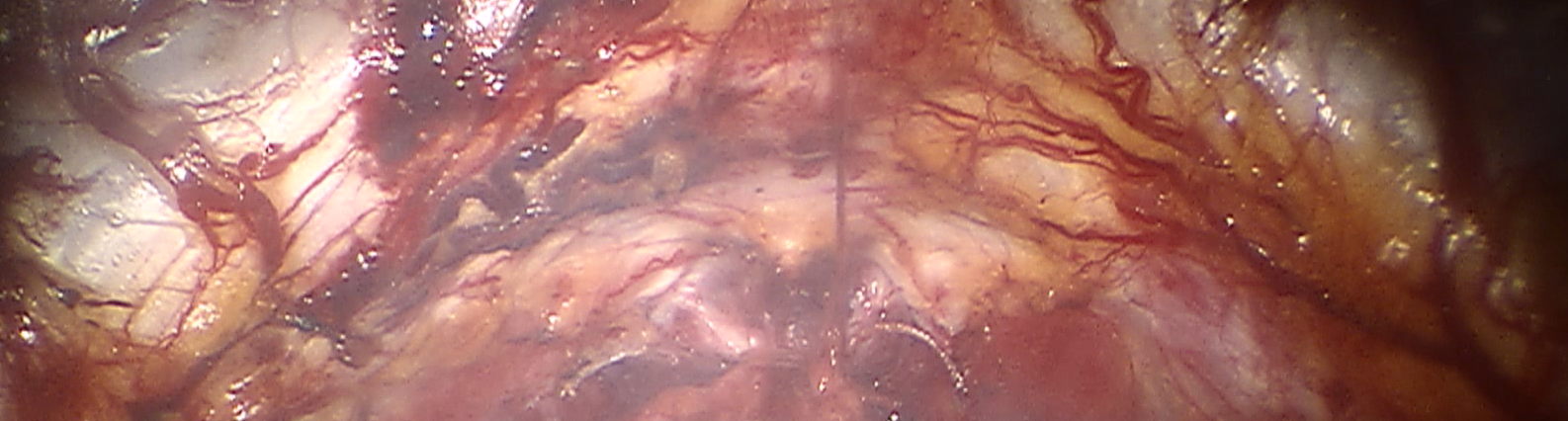}
            \centerline{ (a) Input}
        \end{minipage} &
        \begin{minipage}[b]{0.16\linewidth}\centering
            \includegraphics[width=\linewidth,height=\linewidth]{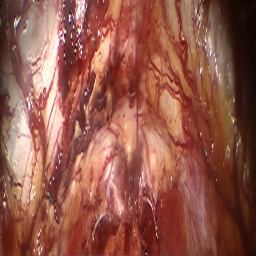}
            \centerline{ (b) case1}
        \end{minipage} &
        \begin{minipage}[b]{0.16\linewidth}\centering
            \includegraphics[width=\linewidth,height=\linewidth]{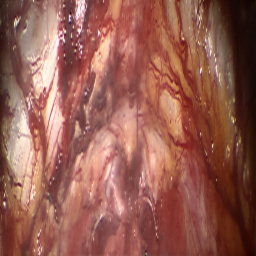}
            \centerline{ (c) case2}
        \end{minipage} &
        \begin{minipage}[b]{0.16\linewidth}\centering
            \includegraphics[width=\linewidth,height=\linewidth]{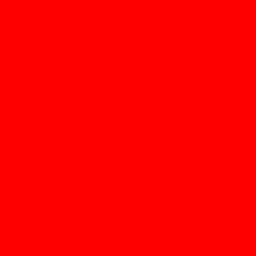}
            \centerline{ (d) case3}
        \end{minipage} &
        \begin{minipage}[b]{0.16\linewidth}\centering
            \includegraphics[width=\linewidth,height=\linewidth]{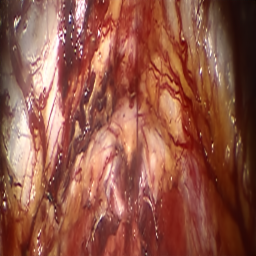}
            \centerline{ (e) Ours}
        \end{minipage} &
        \begin{minipage}[b]{0.16\linewidth}\centering
            \includegraphics[width=\linewidth,height=\linewidth]{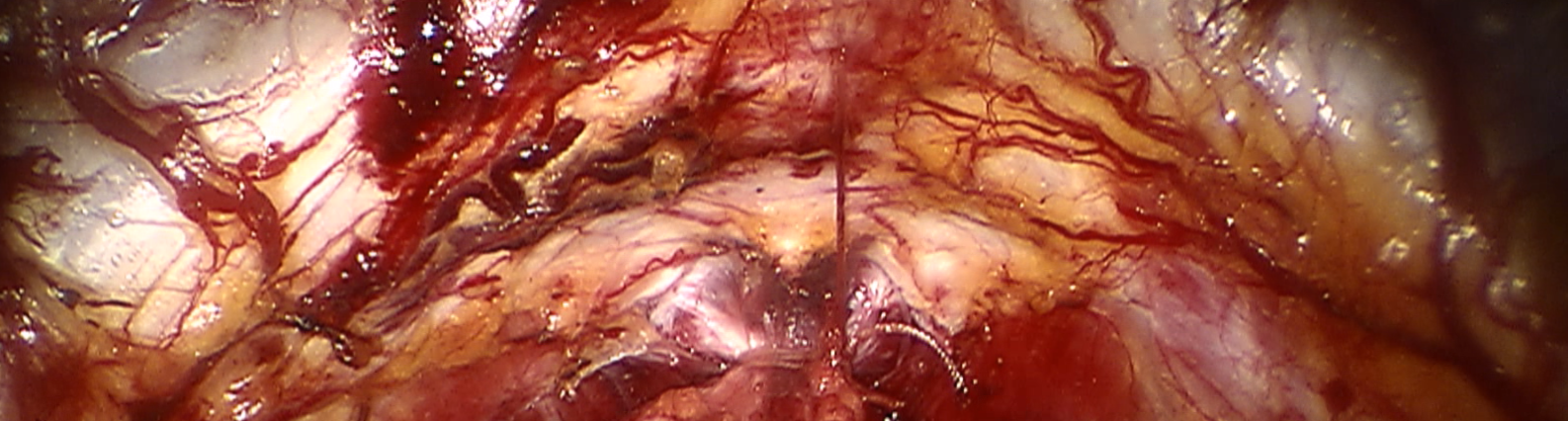}
            \centerline{ (f) Target}
        \end{minipage} \\

    \end{tabular}
    \caption{Ablation comparisons sampled from the DesmokeData and LSD3K datasets. The first two rows are from DesmokeData and the last two rows are from LSD3K. (settings: case1 = ADA+CrossGating, case2 = DHA+CrossGating, case3 = DHA+ADA).}

    \label{fig:ablation}
\end{figure*}

\subsection{Ablation Studies}

To validate the effectiveness and individual contributions of the core components of our proposed RGA-Net, we conducted a series of ablation experiments on the DesmokeData dataset. We systematically analyzed the impact of our key designs: the DHA module, the ADA module, and the Cross-Gating (CG) mechanism for feature fusion. In these studies, we created several variants of our network by removing or replacing one component at a time and evaluated their performance. The quantitative results of these experiments are presented in \cref{tab:ab}, and qualitative visual comparisons are shown in \cref{fig:ablation}.

\subsubsection{Effectiveness of the Dual-Stream Hybrid Attention Module} 
The DHA module is designed to capture both local surgical details and global illumination changes by combining shifted window attention with a frequency-domain processing branch. To ablate its effect, we replaced the DHA modules in the encoder with standard Swin Transformer blocks, thus removing the spectral pathway and the hybrid attention mechanism. As demonstrated in Table II, this variant experienced a significant drop in performance. This decline highlights the importance of integrating frequency-domain information, which is crucial for handling the complex light scattering caused by surgical smoke and restoring high-frequency textural details. The results confirm that the dual-stream approach provides a richer feature representation than spatial attention alone.

\subsubsection{Effectiveness of the Axis-Decomposed Attention Module} 
We then investigated the contribution of the ADA module, which is employed in the decoder and latent space to efficiently process multi-scale features. We created a variant where the ADA modules were substituted with a more conventional self-attention mechanism without the block and grid axis decomposition. The results in Table II show a clear degradation in performance for this variant compared to the full model. This outcome validates that factorizing attention along two distinct axes allows the model to capture both fine-grained local patterns and long-range spatial dependencies more effectively and efficiently. The ADA module's design provides a powerful yet computationally manageable way to model complex feature relationships during the reconstruction phase.

\subsubsection{Effectiveness of the Cross-Gating Mechanism} 
Finally, we analyzed the efficacy of the cross-gating mechanism used for multi-scale feature fusion between the encoder and decoder. We replaced our CG blocks with a standard skip-connection method, specifically simple concatenation followed by a convolutional layer, as is common in many U-Net-based architectures. As shown in Table II, this change resulted in a substantial performance decrease. This finding underscores the superiority of the bidirectional feature modulation offered by our CG mechanism. Unlike simple fusion, cross-gating allows the encoder and decoder pathways to selectively amplify relevant features and suppress irrelevant information from each other, leading to a more refined and effective integration of multi-scale context, which is vital for high-quality image reconstruction.

In conclusion, the ablation studies comprehensively demonstrate that each of our proposed components—DHA, ADA, and CG—is integral to the overall performance of RGA-Net. The full model consistently outperforms all ablated variants, proving the synergistic benefits of our architectural design for the challenging task of surgical smoke removal.

\section{CONCLUSIONS}

This paper presents RGA-Net, a novel deep learning framework for smoke removal in robotic surgery. RGA-Net addresses surgical smoke’s unique challenges, restoring high-quality visual feedback. Key innovations include the DHA module for capturing textures and illumination, the ADA module for multi-scale feature modeling, and the Cross-Gating (CG) mechanism for enhanced feature fusion.

Extensive experiments on the DesmokeData and LSD3K datasets have demonstrated the superiority of our method. RGA-Net consistently outperformed a wide range of state-of-the-art methods in both quantitative metrics and qualitative assessments. The results show that our model not only removes dense and non-homogeneous smoke more effectively but also excels at preserving crucial fine-grained details and maintaining natural color fidelity, which are often lost by other methods. Furthermore, our comprehensive ablation studies have rigorously validated the individual contribution of each proposed module, confirming that the synergistic interplay between DHA, ADA, and CG is key to the network's exceptional performance.

By providing a consistently clear operative field, RGA-Net represents a significant advancement toward a more reliable and safer human-robot interface in surgery. This enhanced visualization has the potential to reduce surgeons' cognitive load, decrease operation times, and ultimately minimize the risk of iatrogenic injury. We believe the proposed framework is a substantial step forward in computational vision enhancement and holds great promise for integration into next-generation robotic surgical systems.






\bibliographystyle{IEEEtran}
\bibliography{ref}

\end{document}